\documentclass[11pt]{article}

\usepackage[]{acl}
\usepackage{amssymb}
\usepackage{amsmath,amssymb}
\usepackage{dsfont}
\usepackage{cuted}
\usepackage{multirow}
\usepackage{array}
\usepackage{microtype}
\usepackage{graphicx}
\usepackage{hyperref}
\usepackage[capitalize]{cleveref}
\crefname{figure}{Fig.}{Figs.}

\usepackage{subfig}
\usepackage{makecell}

\title{Mixture of Heterogeneous Grouped Experts for Language Modeling}

\author{Zhicheng Ma$^{1,2,*}$, Xiang Liu$^{1,2,*}$, Zhaoxiang Liu$^{1,2,*,\dagger}$, Ning Wang$^{1,2}$,\\ Yi Shen$^{1,2}$, Kai Wang$^{1,2}$, Shuming Shi$^{1,2}$ \and Shiguo Lian$^{1,2,\dagger}$ \\
\\
        $^{1}$Data Science \& Artificial Intelligence Research Institute, China Unicom \\ $^{2}$Unicom Data Intelligence, China Unicom\\
        $^{*}$Equal contribution\\ 
        $^{\dagger}$Corresponding authors}

\begin{document}

\maketitle
\begin{abstract}
Large Language Models (LLMs) based on Mixture-of-Experts (MoE) are pivotal in industrial applications for their ability to scale performance efficiently. However, standard MoEs enforce uniform expert sizes, creating a rigidity that fails to align computational costs with varying token-level complexity. While heterogeneous expert architectures attempt to address this by diversifying expert sizes, they often suffer from significant system-level challenges, specifically unbalanced GPU utilization and inefficient parameter utilization, which hinder practical deployment.
To bridge the gap between theoretical heterogeneity and robust industrial application, we propose \textbf{Mixture of Heterogeneous Grouped Experts (MoHGE)} which introduces a two-level routing mechanism to enable flexible, resource-aware expert combinations. To optimize inference efficiency, we propose a \textbf{Group-Wise Auxiliary Loss}, which dynamically steers tokens to the most parameter-efficient expert groups based on task difficulty.
To address the critical deployment challenge of GPU load balancing, we introduce an \textbf{All-size Group-decoupling Allocation strategy} coupled with an \textbf{Intra-Group Experts Auxiliary Loss}. These mechanisms collectively ensure uniform computation distribution across GPUs.
Extensive evaluations demonstrate that MoHGE matches the performance of MoE architectures while reducing the total parameters by approximately 20\% and maintaining balanced GPU utilization. Our work establishes a scalable paradigm for resource-efficient MoE design, offering a practical solution for optimizing inference costs in real-world scenarios. The code is publicly available at https://github.com/UnicomAI/MoHGE. 
\end{abstract}

\section{Introduction}

Transformer-based large language models (LLMs) (\citeauthor{achiam2023gpt}, \citeyear{achiam2023gpt}; \citeauthor{touvron2023llama}, \citeyear{touvron2023llama};
\citeauthor{bai2023qwen}, \citeyear{bai2023qwen};
\citeauthor{liu2024deepseekv2}, \citeyear{liu2024deepseekv3}) have achieved remarkable success across a wide range of natural language processing (NLP) tasks. According to scaling laws (\citeauthor{kaplan2020scaling}, \citeyear{kaplan2020scaling}), larger models consistently deliver better performance, and recent studies (\citeauthor{wei2022emergent}, \citeyear{wei2022emergent}) have shown that scaling can also give rise to emergent abilities. However, the computational cost of training and deploying such large models grows exponentially 
(\citeauthor{thompson2020computational}, \citeyear{thompson2020computational}), creating a critical bottleneck for both research and real-world applications.

Mixture-of-Experts (MoE) architectures, originally proposed in \cite{jacobs1991adaptive} and \cite{jordan1994hierarchical}, offer an effective solution by enabling sparse activation: Only a small subset of the model parameters are engaged in per inference step, allowing the model to scale efficiently without proportionally increasing computational overhead.

Despite this advantage, most existing MoE models consist of experts with identical sizes and structures. This homogeneity poses a limitation when generating tokens of varying difficulty: some tokens are easy to predict, while others require more sophisticated reasoning. To address this, recent approaches such as MoDSE (\citeauthor{sun-etal-2024-mixture}, \citeyear{sun-etal-2024-mixture}) and HMoE (\citeauthor{wang2024hmoe}, \citeyear{wang2024hmoe}) have explored the experts with different sizes.

However, these works still have drawbacks. Specifically, MoDSE employs a routing strategy that promotes uniform routing probabilities across experts, which fail to route input tokens to the most suitable experts, leading to inefficient parameter utilization. In addition, its fully heterogeneous expert design restricts the diversity of expert combinations, resulting in limited performance. HMoE mentions the concept of hybrid heterogeneous–homogeneous experts as a promising direction, but does not explicitly explore this design. Furthermore, this hybrid structure suffers from significant GPU utilization imbalance due to uneven parameter sizes, ultimately degrading training efficiency and limiting its scalability.
\begin{figure}[t]
    \centering
    \includegraphics[width=0.8\linewidth]{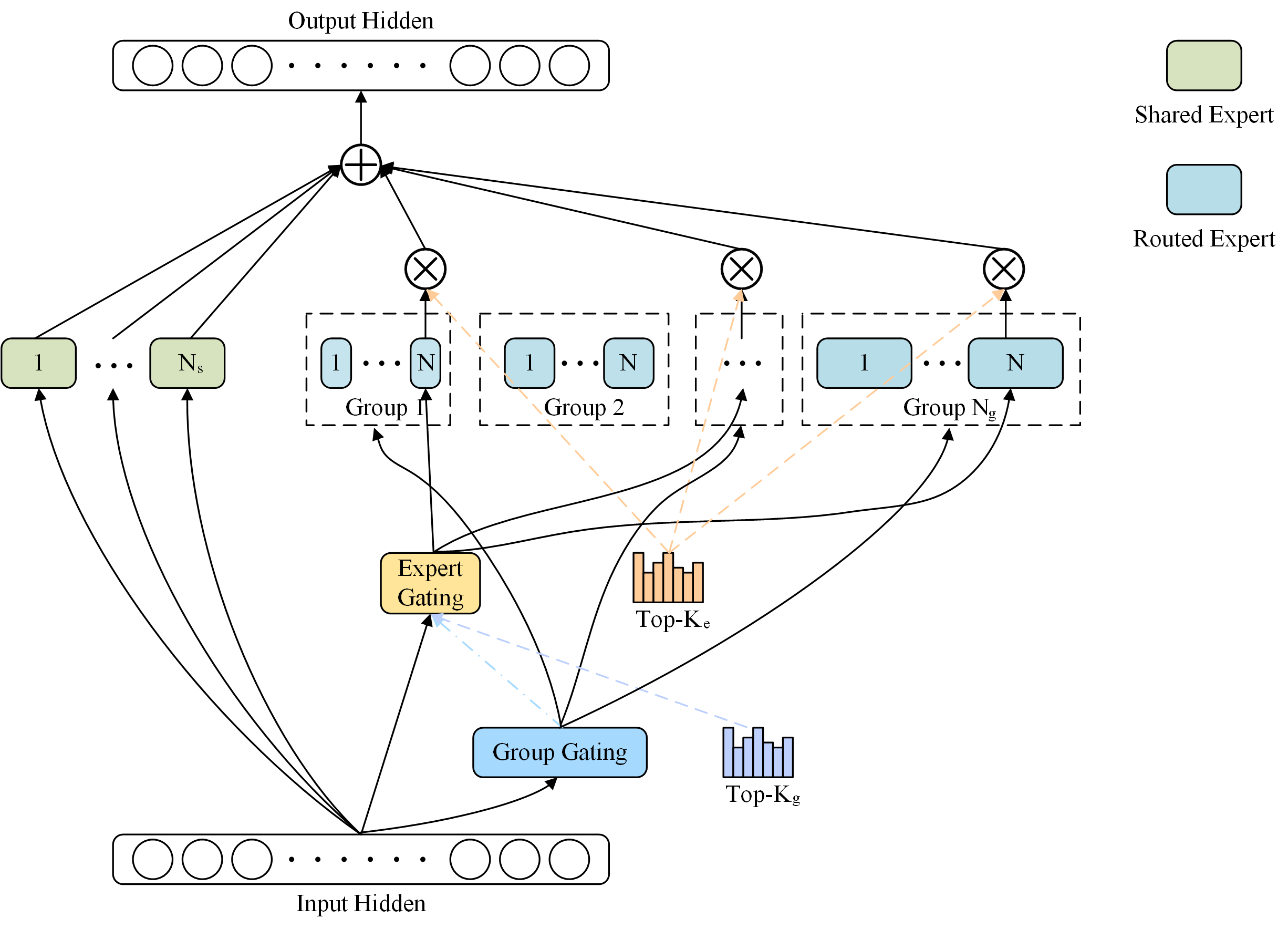}
    \caption{An illustration of our Mixture of Heterogeneous Grouped Experts Layer.}
    \label{fig: moelayer}
    \vspace{-4mm}
\end{figure}

To this end, we propose the Mixture of Heterogeneous Grouped Experts (MoHGE) as shown in \cref{fig: moelayer}.
Experts are organized into multiple groups such that experts within the same group have the same parameter size, while different groups contain experts of different sizes. 
We firstly introduce a two-level routing mechanism which enables more fine-grained and diverse expert combinations.
To adapt model capacity to task complexity, we introduce a Group-Wise Auxiliary Loss that encourages the selection of expert groups with appropriate parameter sizes. 
To mitigate GPU load imbalance, we further propose an All-size Group-decoupling Allocation strategy that evenly distributes experts from each group across all GPUs, ensuring uniform memory usage. In addition, we introduce an Intra-Group Experts Auxiliary Loss that promotes balanced routing among experts within the same group. Together, both strategies lead to more balanced GPU utilization during both training and inference.

Our contributions to efficient industrial LLM are summarized as follows:
\begin{itemize}
\setlength{\itemsep}{0.01em}
\setlength{\parskip}{0.01em}
\item \textbf{Novel Architecture}: We propose a novel MoE architecture, MoHGE, which achieves precise capacity match based on task difficulty and efficient parameter utilization by incorporating the two-level routing strategy and the Group-Wise Auxiliary Loss.
\item \textbf{Load Balance}: 
To ensure balanced GPU utilization, we propose the All-size Group-decoupling Allocation strategy and the Intra-Group Experts Auxiliary Loss. Together, these techniques maintain intra-group utilization equilibrium and achieve uniform GPU workloads, ensuring the model's scalability.
\item \textbf{Empirical Validation}: Experimental results demonstrate that MoHGE achieves an accuracy comparable to that of conventional MoE while reducing total parameters. More noteworthy, detailed routing analysis confirms successful balance of GPU utilization and validates our loss functions' ability to regulate expert activation patterns.
\end{itemize}



\section{Related Work}
Standard MoE architectures \cite{jacobs1991adaptive, shazeer2017outrageously, fedus2022switch} typically rely on homogeneous experts. While effective for scaling, this uniformity creates an efficiency bottleneck: predetermined compute budgets are applied regardless of token complexity. To address this, recent works have explored dynamic compute allocation. \cite{huang-etal-2024-harder} introduced Top-P routing to vary the number of activated experts, yet this approach depends on rigid thresholds and rudimentary difficulty modeling. Others explored heterogeneous expert sizes to match varying task difficulties \cite{sun-etal-2024-mixture, wang2024hmoe}. However, \cite{sun-etal-2024-mixture} suffers from suboptimal routing that misaligns expert size with token needs. While \cite{wang2024hmoe} proposed a hybrid structure, it incurs severe computational imbalance due to uneven expert sizes on GPUs, hindering training efficiency and scalability.

In contrast, our MoHGE architecture introduces a grouped heterogeneous design with a two-level routing mechanism, ensuring precise alignment between model capacity and token difficulty. Crucially, we address the hardware deployment bottlenecks of prior heterogeneous works via our All-size Group-decoupling Allocation, guaranteeing balanced GPU utilization.

\section{Mixture of Heterogeneous Grouped Experts}

\subsection{Group-wise Varied Size Experts}

Traditional MoE architectures typically employ a gating network that routes inputs to a uniform set of experts, all of which have the same model size. However, as shown by \cite{sun-etal-2024-mixture}, the cognitive challenge of predicting the next token varies significantly across different linguistic contexts—mirroring the dynamic processing demands seen in human cognition.

Building on this observation, we introduce a novel heterogeneous expert architecture that organizes experts into multi-granularity groups. Formally, we structure the expert set $\{{E_1, E_2, E_3, \cdots, E_{N_e}}\}$ into distinct groups $\{G_1, G_2, G_3, \cdots, G_{N_g}\}$, where each group contains $N = N_e / N_g$ experts ($N_e$ and $N_g$ denote the total number of experts and groups, respectively).
For the convenience of expression, we transform experts from $E_j$ into $E_{g,i}$, where $g$ represents the group to which the expert belongs and $i$ represents the index of the expert in the group.
Experts within each group share identical parameter sizes, while parameter scales vary across groups according to a predefined progression. Specifically, the hidden dimension of experts in group $G_i$ is given by: \par
\begin{scriptsize}
\begin{equation}
\begin{aligned}
    W_i = 2*W_{\text{base}} - W_{N_g - i}
\end{aligned}
\end{equation}
\end{scriptsize}
where $W_{\text{base}}$ represents the base hidden dimension and the $W_i$ increases as $i$ increases.
This hierarchical organization enables dynamic computation allocation: compact experts efficiently process simpler linguistic patterns, while progressively larger experts with greater capacity handle more complex contextual relationships. 

\subsection{Two-level Routing Mechanism}
To efficiently manage the hierarchical structure of experts, our two-level routing mechanism operates in two stages. The \textbf{group gating model} first selects expert groups based on their relevance to the input, and the \textbf{expert gating model} then chooses specific experts within these groups. This design ensures the computation is focused on the most relevant experts, reducing overhead by restricting selection to the top-$K_g$ groups. Combined with our heterogeneous grouped experts design, this routing mechanism enables a richer and more flexible set of expert combinations, guaranteeing better performance.\par
\subsubsection{Group Gating Model}
The group gating model computes scores $GS$ for all $N_g$ expert groups.
For the $t$-th token input $\mathbf{x}_t$, the score for the $g$-th group is,
\begin{scriptsize}
\vspace{-1mm}
\begin{align}
    GS_{g,t} = Sigmoid({\mathbf{x}_t}^T \mathbf{e}_g)
\end{align}
\vspace{-1mm}
\end{scriptsize}
where $\mathbf{e}_g$ is the centroid embedding of the 
$g$-th expert group. The model then selects the $K_g$
groups with the highest scores, restricting the expert gating model to only route tokens to experts within these groups.

\subsubsection{Expert Gating Model}
The expert gating model operates in three phases: \textbf{Intra-Group Expert Scores Calculation}, \textbf{Experts for Global Selection} and \textbf{Global Normalization}.

\textbf{1. Intra-Group Expert Scores Calculation.} 
For each selected group, the model computes unnormalized scores for its experts using a group-wise Softmax: \par
\begin{scriptsize}
\begin{equation}
\begin{aligned}
ES'_{g,i,t} =  \begin{cases}
\text{Softmax}(\mathbf{x_t}^T \mathbf{e}_{g,i})\ ,  \qquad \text{if } GS_{g,t} \in \text{top}K_g(GS_t) \\
0\ , \qquad \qquad \qquad  \qquad  \text{otherwise}
\end{cases}
\end{aligned}
\end{equation}
\end{scriptsize}
where ${e}_{g,i}$ is the embedding of the $i$-th expert in group $g$.

\textbf{2. Experts for Global Selection.}
The intra-group expert scores are scaled by the group scores to reflect group importance: \par
\begin{scriptsize}
\begin{equation}
{ES''_{g,i,t}}  = (ES' \cdot GS)_{g,i,t}  
\end{equation}
\end{scriptsize}
Next, the model selects the top-$K_e$ experts globally. Scores for all other experts are set to zero: \par
\begin{scriptsize}
\begin{equation}
\begin{aligned}
ES'''_{g,i,t}  = \begin{cases}
    ES''_{g,i,t} \ , \qquad \text{if } {ES''_{g,i,t}} \in \text{top}K_e({ES''_{g,i,t}}) \\
    0\ , \qquad \qquad \qquad  \qquad \text{otherwise}
\end{cases}
\end{aligned}
\end{equation}
\end{scriptsize}

\textbf{3. Global Normalization.}
Finally, the selected expert scores are normalized: \par
\begin{scriptsize}
\begin{equation}
    ES_{g,i,t} = \frac{ES'''_{g,i,t}}{\sum_j^{N_g}\sum_k^{N} ES'''_{j,k,t}}
\end{equation}
\end{scriptsize}

This three-step gating strategy enables fine-grained, efficient expert selection by prioritizing both group relevance and individual expert utility.


\begin{figure}[t]
    \centering
    \includegraphics[width=0.75\linewidth]{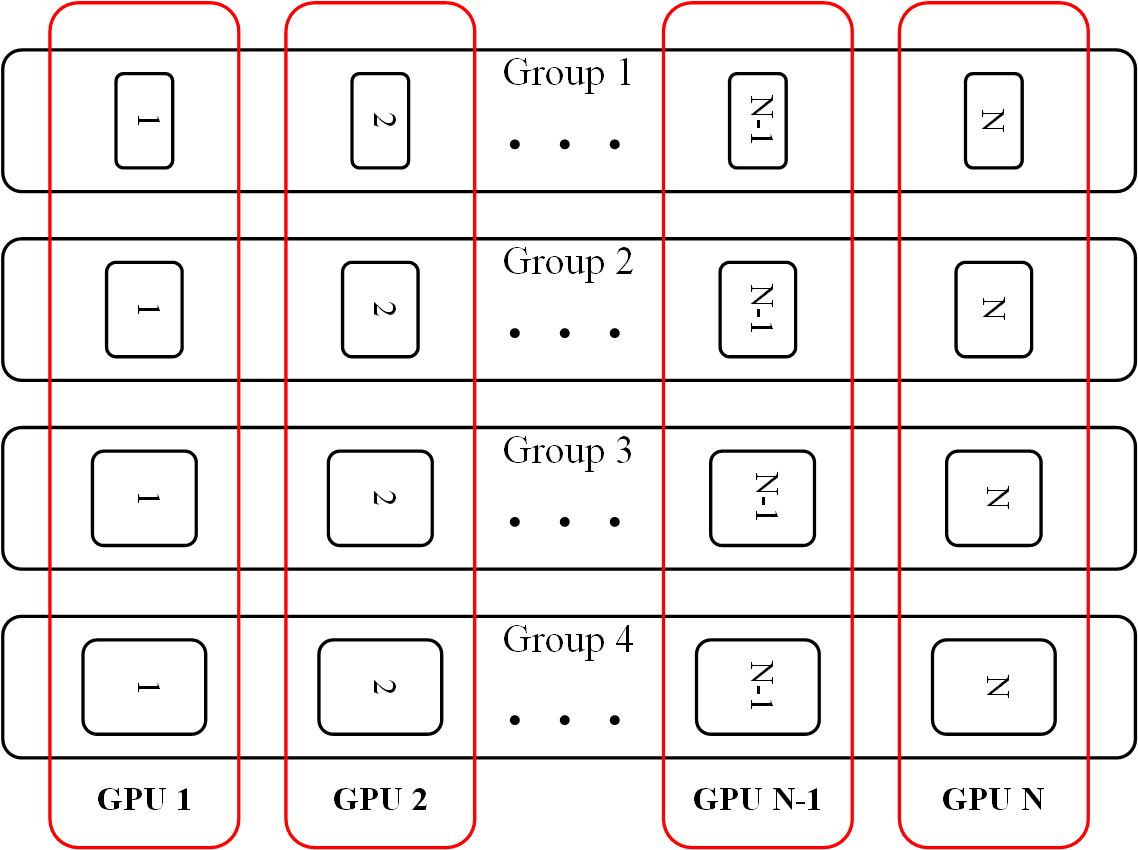}
    \caption{An example of All-size Group-decoupling Allocation.}
    \label{fig:All-size Group-decoupling Allocation}
    \vspace{-4mm}
\end{figure}

\subsection{Efficient Parameter Utilization}
Without regularization, experts with larger parameter sizes tend to dominate the routing decisions due to their stronger representational capacity. This dominance can result in inefficient expert usage, as smaller expert groups with fewer parameters may not be fully utilized. To address this issue and improve parameter utilization, we introduce a slight penalty for expert groups with larger parameter sizes. Specifically, we propose \textbf{Group-Wise Auxiliary Loss} $L_{G}$, which slightly penalizes expert groups with larger parameter sizes.

This loss encourages the gating model to consider groups with fewer parameters, leading to more efficient parameter utilization. The model ultimately learns to trade off between minimizing cross-entropy and reducing parameter-related costs.
The loss is formulated as: \par
\vspace{-2mm}
\begin{scriptsize}
\begin{align}
    L_{G} &= \alpha_{G}\sum_{i=1}^{N_g}\frac{W_i}{W_{max}}f^{G}_{i} p^{G}_{i} \\
    f^{G}_{i} &= \frac{N_g}{K_g}\sum_{t=1}^{T}\mathds{1}(GS_{i,t} \in topk(GS_t)) \\
    p^{G}_{i} &= \frac{1}{T}\sum_i^{T}{s^{G}_{i,t}}' \\
    {s^{G}_{i,t}}' &= \frac{GS_{i,t}}{\sum_j^{N_g}GS_{j,t}}  
\end{align}
\end{scriptsize}
where $W_i$ is the parameter count of group $i$, $f^{G}_{i}$ is the group’s routing frequency, balance factor $\alpha_{G}$ is assigned an extremely small value and $p^{G}_{i}$ is its average normalized routing score.

\subsection{Load Balance Consideration}

Experts with larger hidden dimensions (i.e., those exceeding a base width $W_\text{base}$) introduce disproportionately higher memory and computational costs. If not carefully managed, this imbalance can lead to severe GPU load imbalances, where certain GPUs become bottlenecks while others remain underutilized. This inefficiency hampers overall training performance and scalability. To mitigate this issue, we introduce \textbf{All-size Group-decoupling Allocation} and \textbf{Intra-Group Experts Auxiliary Loss}, which work synergistically to achieve a uniform distribution of computational load across GPUs, thus ensuring balanced resource utilization.

\subsubsection{Allocation Strategy}

An All-size expert set consists of the $i$-th expert from all groups. Each GPU is assigned multiple such sets, ensuring that the total number of expert parameters on each GPU remains consistent and smoothing out the variance in parameter size across the system.
If expert workloads are evenly balanced within each group (which is encouraged by our auxiliary loss design), this approach leads to balanced GPU utilization overall.

As illustrated in \cref{fig:All-size Group-decoupling Allocation} (with $N_g = 4$), each GPU hosts one All-size expert set (e.g., experts ${E_{1,i}, E_{2,i}, E_{3,i}, E_{4,i}}$). Regardless of the group selection during routing, as long as expert activation within each group is balanced, overall GPU resource usage remains evenly distributed.
\begin{table*}[!htb]
\centering
\scriptsize
\setlength{\tabcolsep}{6.25pt}
\begin{tabular}{c|cc|ccccccc}
\hline
Method     & \begin{tabular}[c]{@{}c@{}}Total \\ Parameters\end{tabular} & \begin{tabular}[c]{@{}c@{}@{}}Activated \\ Parameters \\ of Experts\end{tabular} & MMLU  & SIQA  & GSM8K & LAMBADA & MATH & PIQA  & TriviaQA \\  \hline

Dense      & 0.807B                                                      & \---                                                         &  26.36     & 35.30      &   2.79    &   61.02      &   1.33   &  47.35     &    34.98      \\ 
MoE-3B     & 3.3614B                                                     & 0.376B                                                         & 26.22      &  35.41     &       3.03&         60.86&      1.34&       \textbf{49.08} &          39.16\\
MoHGE-3B & 2.821B                                                      & 0.295B                                                         & \textbf{26.41}      & \textbf{35.57}      &       \textbf{4.02} &         \textbf{62.37}&      \textbf{1.36}&       \textbf{49.08} &          \textbf{39.20} \\  \hline

Dense      & 1.672B                                                      & \---                                                         &  30.78     & 42.29      &   4.62    &   68.05      &   6.82   &  54.92     &    50.26      \\ 
MoE-14B     & 16.760B                                                     & 1.191B                                                         & 31.18      &  44.28     &  4.92    &  67.94  & 7.30  &       56.71    &  51.77  \\
MoHGE-14B & 14.122B                                                      & 0.843B                                                         & \textbf{31.62}      & \textbf{45.62}      &      \textbf{5.76} &\textbf{69.89}   & \textbf{9.42}    &\textbf{58.73} &\textbf{52.69} \\  \hline
\end{tabular}
\caption{Comparison between Dense model, MoE baseline and our MoHGE, the highest scores for each benchmark is highlighted in bold.}
\label{tab:evaluations}
\vspace{-3mm}
\end{table*}

\begin{table}[!htb]
\centering
\scriptsize
\scriptsize
\setlength{\tabcolsep}{2pt}
\begin{tabular}{l|cc|cc|cc}
\hline
\multicolumn{1}{c|}{\multirow{2}{*}{Benchmark}} & \multicolumn{2}{c|}{Dense-Model} & \multicolumn{2}{c|}{MoE} & \multicolumn{2}{c}{MoHGE} \\ \cline{2-7}
\multicolumn{1}{c|}{}           & 0.81B    & 1.67B     & 3B     & 14B     & 3B     & 14B     \\ \hline
MMLU                            & 9.06h    & 17.83h      & 9.85h  & 19.27h   & 9.58h  & 18.86h  \\
SIQA                           & 1.09h    & 2.17h    & 1.29h  & 2.51h   & 1.17h  & 2.33h   \\
GSM8K                          & 0.77h    & 1.52h    & 0.84h  & 1.63h     & 0.86h  & 1.62h   \\
LAMBADA                         & 2.83h    & 5.70h    & 3.17h  & 6.27h   & 3.03h  & 6.08h   \\
MATH                          & 2.86h    & 5.74h     & 3.18h  & 6.33h    & 3.01h  & 6.12h   \\
PIQA                           & 1.03h    & 2.07h     & 1.20h  & 2.38h  & 1.09h  & 2.17h    \\
TriviaQA                        & 5.24h    & 10.39h    & 5.78h  & 11.46h   & 5.46h  & 10.85h   \\ \hline
\end{tabular}
\caption{The inference duration of the Dense, MoE and MoHGE models on downstream tasks.}
\label{tab: inference time}
\vspace{-3mm}
\end{table}

\subsubsection{Intra-Group Experts Auxiliary Loss}
In addition to the standard cross-entropy loss, we incorporate an intra-group experts auxiliary loss $L_{E}$ adapted from DeepSeekV2 (\citeauthor{liu2024deepseekv2}, \citeyear{liu2024deepseekv3}) to encourage balanced expert usage during routing. While DeepSeekV2 penalizes imbalance across all experts globally, our approach focuses on experts within each selected group, promoting uniform routing frequencies locally.
This design ensures that all experts within an active group are selected with equal frequency during training, leading to better load distribution across GPUs. The auxiliary loss is defined as: 

\begin{scriptsize}
\begin{align}
    &L_{E} = \alpha_{E}\sum_{g=1}^{N_g}\sum_{i=1}^{N}f^{E}_{g,i} p^{E}_{g,i} \\
    &f^{E}_{g,i} = \frac{N}{K_e}\sum_{t=1}^{T}\mathds{1}(ES'_{g,i,t} \in topK_e(ES'_t)) \\
    &p^{E}_{g,i} = \frac{1}{T}\sum_i^{T} S_{g,j,t}^{E} \\
    &S_{g,j,t}^{E} = \frac{ES'_{g,i,t}}{\sum_j^{N}ES'_{g,j,t} + \epsilon}
\end{align}
\end{scriptsize}

\noindent where $f^{E}_{g,i}$ represents the normalized routing frequency of the $i$-th expert in group $g$, ${S^{E}_{g,i,t}}$ is the normalized routing score, $p^{E}_{g,i}$ is the average selection probability across time steps, the balance factor $\alpha_{E}$ is assigned an extremely small value and $\epsilon$ is a very small constant to ensure that the denominator is not 0.

\section{Experiments}

\subsection{Main Results}

Following OpenCompass protocols (\citeauthor{contributors2023opencompass}, \citeyear{contributors2023opencompass}), \cref{tab:evaluations} reports the zero-shot or few-shot (\citeauthor{kojima2022large}, \citeyear{kojima2022large}; \citeauthor{brown2020language}, \citeyear{brown2020language}) in-context learning performance of our pretrained MoHGE models on a diverse suite of downstream tasks, including MMLU (\citeauthor{hendrycks2020measuring}, \citeyear{hendrycks2020measuring}), SIQA (\citeauthor{sap2019socialiqa}, \citeyear{sap2019socialiqa}), GSM8K (\citeauthor{cobbe2021training}, \citeyear{cobbe2021training}), LAMBADA (\citeauthor{paperno2016lambada}, \citeyear{paperno2016lambada}), MATH (\citeauthor{hendrycks2024measuring}, \citeyear{hendrycks2024measuring}), PIQA (\citeauthor{bisk2020piqa}, \citeyear{bisk2020piqa}) \cite{bisk2020piqa} and TriviaQA (\citeauthor{joshi2017triviaqa}, \citeyear{joshi2017triviaqa}).

\begin{table*}[t]
\centering
\scriptsize
\setlength{\tabcolsep}{9.5pt}
\begin{tabular}{c|cccccccc|c}
\hline
& GPU\_1 & GPU\_2 & GPU\_3 & GPU\_4 & GPU\_5 & GPU\_6 & GPU\_7 & GPU\_8  & Std      \\ \hline
Group 1	& 12.6\%	& 12.7\%	& 12.8\%	& 11.9\%	& 12.4\%	& 12.3\%	& 12.5\%	& 12.8\%        & 0.00302  \\
Group 2	& 12.7\%	& 12.4\%	& 12.4\%	& 12.9\%	& 12.2\%	& 12.5\%	& 12.4\%	& 12.3\%        & 0.0023   \\
Group 3	& 12.2\%	& 12.6\%	& 12.4\%	& 12.8\%	& 12.3\%	& 12.7\%	& 12.5\%	& 12.5\%        & 0.0020   \\
Group 4	& 12.0\%	& 13.0\%	& 12.5\%	& 12.6\%	& 12.5\%	& 12.4\%	& 12.8\%	& 12.3\%       & 0.00304    \\
Group 5	& 12.6\%	& 12.2\%	& 12.7\%	& 12.8\%	& 12.7\%	& 12.3\%	& 12.5\%	& 12.2\%        & 0.0024   \\
Group 6	& 12.6\%	& 12.6\%	& 12.5\%	& 12.0\%	& 12.2\%	& 12.7\%	& 12.7\%	& 12.7\%        & 0.00302   \\
Group 7	& 12.1\%	& 12.4\%	& 12.7\%	& 12.5\%	& 12.5\%	& 12.6\%	& 12.8\%	& 12.4\%       & 0.0021   \\
Group 8	& 12.9\%	& 12.3\%	& 12.3\%	& 12.4\%	& 12.0\%	& 12.8\%	& 12.5\%	& 12.7\%        & 0.0029  \\ \hline
\end{tabular}
\caption{For 14B scale, number of tokens routed to each GPU roughly closes to the average value.}
\label{tab: token routed}
\vspace{-2mm}
\end{table*}

\begin{table*}[!htb]
\centering
\scriptsize
\setlength{\tabcolsep}{9.5pt}
\begin{tabular}{ccccccccc}
\hline
Token Ranks & Group 1 & Group 2 & Group 3 & Group 4 & Group 5 & Group 6 & Group 7 & Group 8 \\ \hline
Top 1K              & 16.3\%           &  14.9\%           &  14.1\%          &  12.2\%           &  12.3\%           &  10.5\%           & 10.0\%          &  9.7\%            \\
Top 1K-5K            &  15.0\%           &  14.4\%           & 13.4\%         & 13.1\%           & 12.4\%           &  10.3\%          &  10.9\%           & 10.5\%           \\
Top 5K-10K           &  13.6\%           & 13.4\%           &  13.7\%         &  12.6\%          &  11.5\%           &  12.4\%           &  11.5\%          &  11.3\%           \\
Beyond 10K                                                        & 11.3\%           & 12.0\%           &  11.4\%           &  12.7\%          &  13.0\%           &  12.7\%           &  13.6\%           &  13.3\%  \\
\hline
\end{tabular}
\caption{Ratios of tokens with different difficulty (Token Ranks) routed to different expert groups.}
\label{tab: rationofdifficulty1}
\end{table*}

\begin{table*}[!htb]
\centering
\scriptsize
\setlength{\tabcolsep}{9.5pt}
\begin{tabular}{ccccccccc}
\hline
Perplexity & Group 1 & Group 2 & Group 3 & Group 4 & Group 5 & Group 6 & Group 7 & Group 8 \\ \hline
$<=5$ & 14.0\% & 13.3\% & 13.9\% & 13.4\% & 12.6\% & 11.2\% & 11.7\% & 10.0\% \\
$<=10$ & 13.1\% & 13.3\% & 13.1\% & 12.2\% & 12.5\% & 11.9\% & 11.5\% & 12.3\% \\
$>10$ & 12.2\% & 11.6\% & 12.7\% & 11.4\% & 12.8\% & 12.9\% & 13.2\% & 13.2\%   \\
\hline
\end{tabular}
\caption{Ratios of tokens with different difficulty (Perplexity) routed to different expert groups.}
\label{tab: rationofdifficulty2}
\end{table*}

\begin{table*}[!htb]
\centering
\scriptsize
\setlength{\tabcolsep}{8pt}
\begin{tabular}{l|ccccccc|c}
\hline
Method   & MMLU  & SIQA  & GSM8K & LAMBADA & MATH & PIQA  & TriviaQA & GPU Utilization \\ \hline
MoHGE-3B & \textbf{26.41} & 35.57 & \textbf{4.02}  & \textbf{62.37}   & \textbf{1.36} & \textbf{49.08} & \textbf{39.20}    & \textbf{balanced}         \\
MoDSE-3B & 26.27 & 35.47 & 3.71  & 61.98   & 1.32 & 48.91 & 39.16    & \textbf{balanced}          \\
HMoE-3B  & 26.34 & \textbf{35.62} & 3.94  & 62.25   & 1.34 & 49.02 & 39.18    & unbalanced   \\  \hline   
\end{tabular}
\caption{Comparison with MoDSE and HMoE at 3B scale.}
\label{tab:comMoDSEandHMoE}
\vspace{-3.5mm}
\end{table*}

As reported in \cref{tab:evaluations}, averaged over three evaluate runs, MoHGE consistently outperforms both conventional MoE and dense models across all scales. With approximately 20\% fewer parameters, MoHGE achieves comparable or better performance than standard MoE baselines. Compared to the MoE baseline, MoHGE achieves a more favorable trade-off between parameter efficiency and downstream performance by activating fewer expert parameters while simultaneously requiring fewer total parameters.

Specifically, MoHGE reduces the overall parameter count by nearly 20\% relative to standard MoE, and the number of activated parameters in the expert layer is reduced by approximately one quarter. This substantial reduction highlights its effectiveness in balancing model capacity with efficiency.

\begin{figure}[t]
\centering
\subfloat[]{\includegraphics[width=0.25\textwidth]{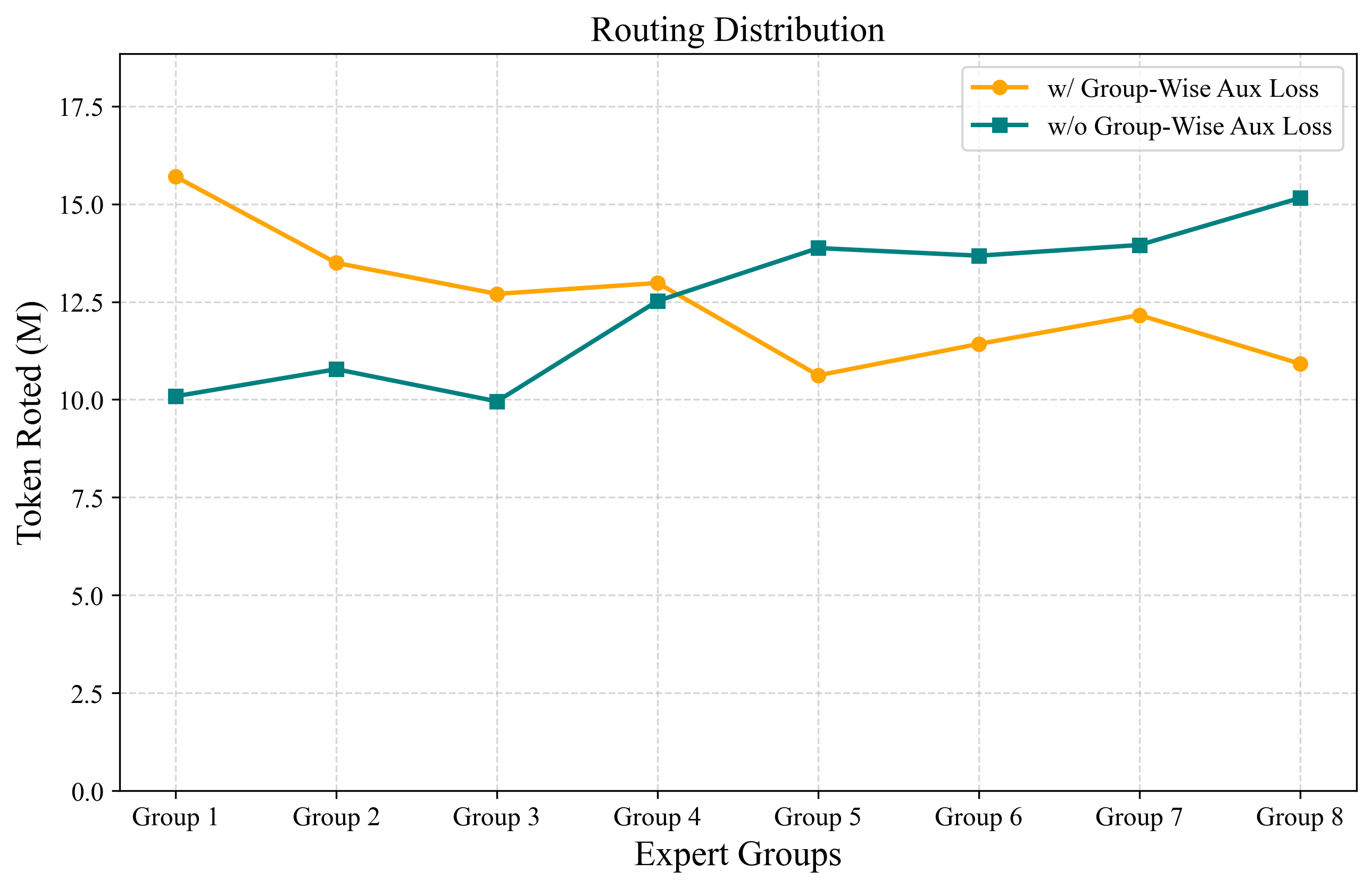}}
\subfloat[]{\includegraphics[width=0.25\textwidth]{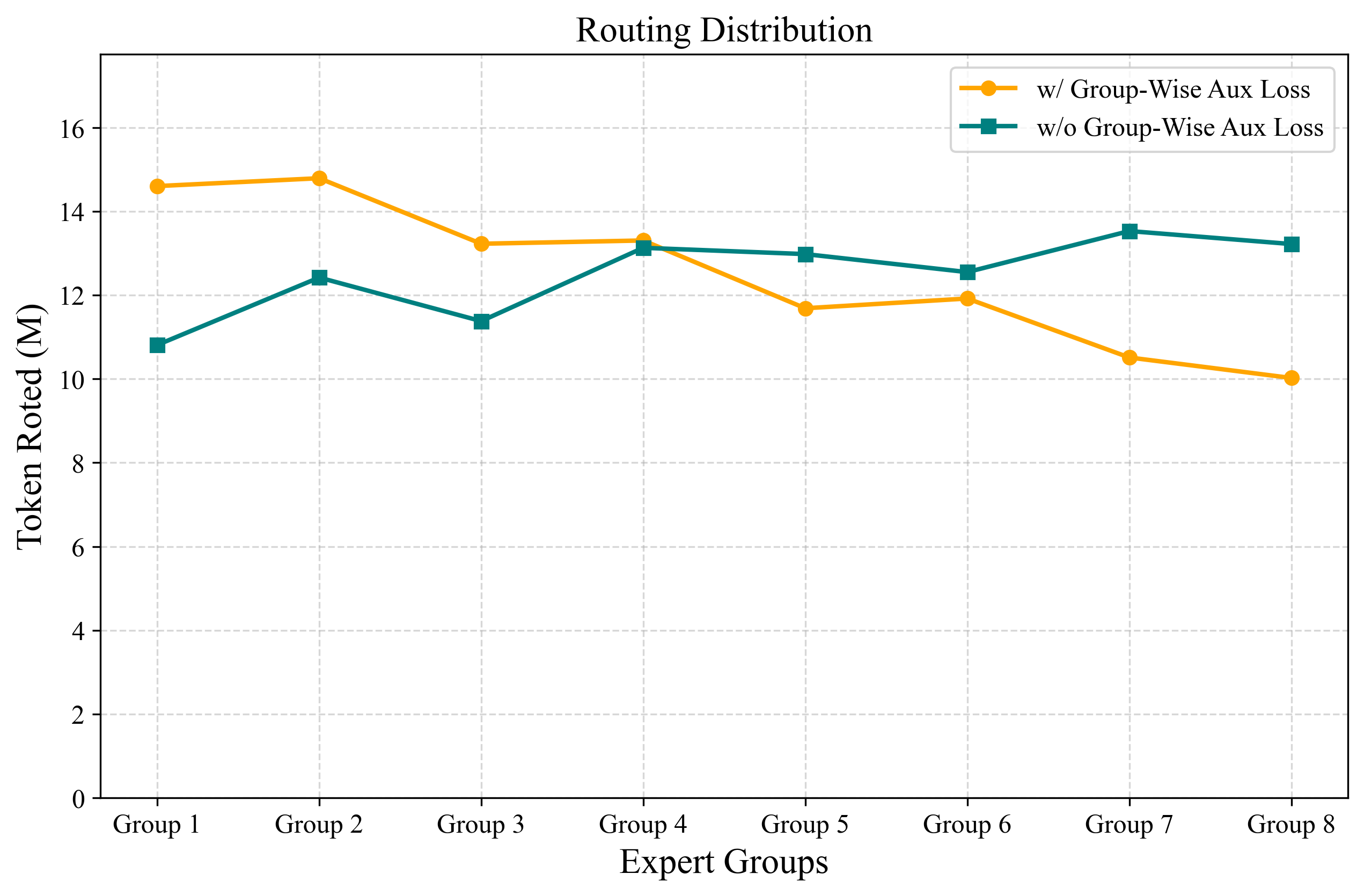}} 
\caption{The number of tokens routed to each expert group. (a) MoHGE-3B \textit{w/} Group-Wise Auxiliary Loss and \textit{w/o} Group-Wise Auxiliary Loss. (b) MoHGE-14B \textit{w/} Group-Wise Auxiliary Loss and \textit{w/o} Group-Wise Auxiliary Loss.}
\label{fig: token routing}
\vspace{-3mm}
\end{figure}

The inference times are demonstrated in \cref{tab: inference time}.
Regarding the slight increase in inference time on GSM8K which is a complex mathematical reasoning task, our routing analysis reveals that the MoHGE tends to select expert groups with larger parameter on GSM8K and this achieves higher accuracy while resulting in more inference time.
Altogether, our model achieves relatively faster inference speeds, showing superior inference efficiency. 

It is also important to note that the dense baseline exhibits faster inference in some cases, as its total parameters are approximately equal to the number of activated parameters in the MoE models. Consequently, dense model can benefit from more streamlined execution without routing overhead. Despite this, MoHGE maintains a favorable balance between inference efficiency and performance by selectively allocating computational resources when needed.

\subsection{Analysis on Token Routing}

\subsubsection{Loss Function}
We conducted a statistical analysis of the distribution of 100 million token routes with and without Group-Wise Auxiliary Loss. As shown in  \cref{fig: token routing}, introducing group routing loss alters token routing behavior: tokens no longer primarily favor larger expert groups, but are instead distributed more widely among smaller expert groups. This indicates that, with a relatively uniform route distribution, the loss encourages the selection of smaller expert groups that can handle the current task difficulty.

\subsubsection{GPU Utilization}
To rigorously evaluate the balancing of GPU utilization, we conduct a GPU-level assessment for 14B scale model by strategically assigning the $i$-th expert from each capacity group to the $i$-th GPU. This experimental design allows us to precisely track how tokens are distributed across experts of varying sizes on each GPU, which reflects the frequency of token processed by experts of different sizes on each GPU.
\cref{tab: token routed} shows that experts of uniform size receive nearly equal routing frequencies across GPUs, indicating balanced intra-group expert and GPU utilization. This confirms that our All-size Group-decoupling Allocation and Intra-group Experts Auxiliary Loss effectively maintain equilibrium in both computational resource loading and expert activation patterns.

\subsubsection{Tokens of Different Difficulties}
We use two methods to classify the difficulty of tokens: one is by occurrence frequency, and the other is by perplexity.
We categorized the vocabulary into four difficulty levels based on occurrence frequency ranks in training corpus: Top 1K (easiest), Top 1K-5K, Top 5K-10K and Beyond 10K (most difficult).
We categorized the vocabulary into three difficulty levels based on perplexity in training corpus: perplexity $<=5$ (easiest), perplexity $<=10$, and perplexity $>10$ (most difficult).
\cref{tab: rationofdifficulty1} and \cref{tab: rationofdifficulty2} show the ratios of tokens with different difficulty routed to different expert groups with MoHGE-3B.
These results demonstrate that simpler tokens tend to be routed to expert groups with fewer parameters, and this validates the effectiveness of our method.

\subsection{Comparison with other heterogeneous MoE}
We reproduced MoDSE and HMoE (Top-P) at the 3B parameter scale and compared them with our MoHGE model. As shown in \cref{tab:comMoDSEandHMoE}, both MoDSE and our MoHGE ensure balanced GPU utilization, while MoHGE achieves consistently better performance across multiple datasets. Although HMoE attains performance comparable to our MoHGE, it can not guarantee balanced GPU utilization, restricting its scalability in larger scale. These results highlight the effectiveness of MoHGE in achieving both strong performance and efficient resource utilization, especially in scalability potential.

The experimental setup and more ablation study can be found in \textbf{Appendix}.

\section{Conclusion}
In this work, we propose MoHGE architecture that introduces group-wise expert size variation to better accommodate the diverse complexity of token predictions. We further design a novel routing mechanism and GPU allocation strategy, combining a new training objective, to guarantee excellent performance, efficient parameter utilization, balanced GPU utilization and better scalability. With approximately 20\% fewer parameters, MoHGE achieves comparable or slightly better performance than standard MoE baselines, and outperforms recent heterogeneous MoE models on most benchmark datasets. By rethinking how expert capacities should vary and be allocated, MoHGE paves the way for developing more efficient and capable large language models.

\section{Ethics Statement}
This work focuses on optimizing the efficiency and deployability of Mixture-of-Experts (MoE) large language models (LLMs) through the proposed Mixture of Heterogeneous Grouped Experts (MoHGE) architecture. We strictly adhere to the ethical guidelines and principles of the Association for Computational Linguistics (ACL) throughout the entire research process.
All co-authors of this work have contributed to the research in compliance with ethical standards, and there are no conflicts of interest to disclose.

\section*{Acknowledgments}
This work was supported by the National Natural Science Foundation of China Enterprise Innovation and Development Joint Fund Project (Grant No. U24B20181).

\bibliography{custom}

\begin{thebibliography}{26}
\providecommand{\natexlab}[1]{#1}

\bibitem[{Achiam et~al.(2023)Achiam, Adler, Agarwal, Ahmad, Akkaya, Aleman,
  Almeida, Altenschmidt, Altman, Anadkat et~al.}]{achiam2023gpt}
Josh Achiam, Steven Adler, Sandhini Agarwal, Lama Ahmad, Ilge Akkaya,
  Florencia~Leoni Aleman, Diogo Almeida, Janko Altenschmidt, Sam Altman,
  Shyamal Anadkat, and 1 others. 2023.
\newblock Gpt-4 technical report.
\newblock \emph{arXiv preprint arXiv:2303.08774}.

\bibitem[{Bai et~al.(2023)Bai, Bai, Chu, Cui, Dang, Deng, Fan, Ge, Han, Huang
  et~al.}]{bai2023qwen}
Jinze Bai, Shuai Bai, Yunfei Chu, Zeyu Cui, Kai Dang, Xiaodong Deng, Yang Fan,
  Wenbin Ge, Yu~Han, Fei Huang, and 1 others. 2023.
\newblock Qwen technical report.
\newblock \emph{arXiv preprint arXiv:2309.16609}.

\bibitem[{Bisk et~al.(2020)Bisk, Zellers, Gao, Choi et~al.}]{bisk2020piqa}
Yonatan Bisk, Rowan Zellers, Jianfeng Gao, Yejin Choi, and 1 others. 2020.
\newblock Piqa: Reasoning about physical commonsense in natural language.
\newblock In \emph{Proceedings of the AAAI conference on artificial
  intelligence}, volume~34, pages 7432--7439.

\bibitem[{Brown et~al.(2020)Brown, Mann, Ryder, Subbiah, Kaplan, Dhariwal,
  Neelakantan, Shyam, Sastry, Askell et~al.}]{brown2020language}
Tom Brown, Benjamin Mann, Nick Ryder, Melanie Subbiah, Jared~D Kaplan, Prafulla
  Dhariwal, Arvind Neelakantan, Pranav Shyam, Girish Sastry, Amanda Askell, and
  1 others. 2020.
\newblock Language models are few-shot learners.
\newblock \emph{Advances in neural information processing systems},
  33:1877--1901.

\bibitem[{Cobbe et~al.(2021)Cobbe, Kosaraju, Bavarian, Chen, Jun, Kaiser,
  Plappert, Tworek, Hilton, Nakano et~al.}]{cobbe2021training}
Karl Cobbe, Vineet Kosaraju, Mohammad Bavarian, Mark Chen, Heewoo Jun, Lukasz
  Kaiser, Matthias Plappert, Jerry Tworek, Jacob Hilton, Reiichiro Nakano, and
  1 others. 2021.
\newblock Training verifiers to solve math word problems.
\newblock \emph{arXiv preprint arXiv:2110.14168}.

\bibitem[{Contributors(2023)}]{contributors2023opencompass}
OpenCompass Contributors. 2023.
\newblock Opencompass: A universal evaluation platform for foundation models.

\bibitem[{Fedus et~al.(2022)Fedus, Zoph, and Shazeer}]{fedus2022switch}
William Fedus, Barret Zoph, and Noam Shazeer. 2022.
\newblock Switch transformers: Scaling to trillion parameter models with simple
  and efficient sparsity.
\newblock \emph{Journal of Machine Learning Research}, 23(120):1--39.

\bibitem[{Hendrycks et~al.(2020)Hendrycks, Burns, Basart, Zou, Mazeika, Song,
  and Steinhardt}]{hendrycks2020measuring}
Dan Hendrycks, Collin Burns, Steven Basart, Andy Zou, Mantas Mazeika, Dawn
  Song, and Jacob Steinhardt. 2020.
\newblock Measuring massive multitask language understanding.
\newblock \emph{arXiv preprint arXiv:2009.03300}.

\bibitem[{Hendrycks et~al.(2024)Hendrycks, Burns, Kadavath, Arora, Basart,
  Tang, Song, and Steinhardt}]{hendrycks2024measuring}
Dan Hendrycks, Collin Burns, Saurav Kadavath, Akul Arora, Steven Basart, Eric
  Tang, Dawn Song, and Jacob Steinhardt. 2024.
\newblock Measuring mathematical problem solving with the math dataset, 2021.
\newblock \emph{URL https://arxiv. org/abs/2103.03874}.

\bibitem[{Huang et~al.(2024)Huang, An, Zhuang, Tao, Zhang, Jin, Xu, Xu, Chen,
  Huang, and Feng}]{huang-etal-2024-harder}
Quzhe Huang, Zhenwei An, Nan Zhuang, Mingxu Tao, Chen Zhang, Yang Jin, Kun Xu,
  Kun Xu, Liwei Chen, Songfang Huang, and Yansong Feng. 2024.
\newblock \href {https://doi.org/10.18653/v1/2024.acl-long.696} {Harder task
  needs more experts: Dynamic routing in {M}o{E} models}.
\newblock In \emph{Proceedings of the 62nd Annual Meeting of the Association
  for Computational Linguistics (Volume 1: Long Papers)}, pages 12883--12895,
  Bangkok, Thailand. Association for Computational Linguistics.

\bibitem[{Jacobs et~al.(1991)Jacobs, Jordan, Nowlan, and
  Hinton}]{jacobs1991adaptive}
Robert~A Jacobs, Michael~I Jordan, Steven~J Nowlan, and Geoffrey~E Hinton.
  1991.
\newblock Adaptive mixtures of local experts.
\newblock \emph{Neural computation}, 3(1):79--87.

\bibitem[{Jordan and Jacobs(1994)}]{jordan1994hierarchical}
Michael~I Jordan and Robert~A Jacobs. 1994.
\newblock Hierarchical mixtures of experts and the em algorithm.
\newblock \emph{Neural computation}, 6(2):181--214.

\bibitem[{Joshi et~al.(2017)Joshi, Choi, Weld, and
  Zettlemoyer}]{joshi2017triviaqa}
Mandar Joshi, Eunsol Choi, Daniel~S Weld, and Luke Zettlemoyer. 2017.
\newblock Triviaqa: A large scale distantly supervised challenge dataset for
  reading comprehension.
\newblock \emph{arXiv preprint arXiv:1705.03551}.

\bibitem[{Kaplan et~al.(2020)Kaplan, McCandlish, Henighan, Brown, Chess, Child,
  Gray, Radford, Wu, and Amodei}]{kaplan2020scaling}
Jared Kaplan, Sam McCandlish, Tom Henighan, Tom~B Brown, Benjamin Chess, Rewon
  Child, Scott Gray, Alec Radford, Jeffrey Wu, and Dario Amodei. 2020.
\newblock Scaling laws for neural language models.
\newblock \emph{arXiv preprint arXiv:2001.08361}.

\bibitem[{Kojima et~al.(2022)Kojima, Gu, Reid, Matsuo, and
  Iwasawa}]{kojima2022large}
Takeshi Kojima, Shixiang~Shane Gu, Machel Reid, Yutaka Matsuo, and Yusuke
  Iwasawa. 2022.
\newblock Large language models are zero-shot reasoners.
\newblock \emph{Advances in neural information processing systems},
  35:22199--22213.

\bibitem[{Liu et~al.(2024{\natexlab{a}})Liu, Feng, Wang, Wang, Liu, Zhao,
  Dengr, Ruan, Dai, Guo et~al.}]{liu2024deepseekv2}
Aixin Liu, Bei Feng, Bin Wang, Bingxuan Wang, Bo~Liu, Chenggang Zhao, Chengqi
  Dengr, Chong Ruan, Damai Dai, Daya Guo, and 1 others. 2024{\natexlab{a}}.
\newblock Deepseek-v2: A strong, economical, and efficient mixture-of-experts
  language model.
\newblock \emph{arXiv preprint arXiv:2405.04434}.

\bibitem[{Liu et~al.(2024{\natexlab{b}})Liu, Feng, Xue, Wang, Wu, Lu, Zhao,
  Deng, Zhang, Ruan et~al.}]{liu2024deepseekv3}
Aixin Liu, Bei Feng, Bing Xue, Bingxuan Wang, Bochao Wu, Chengda Lu, Chenggang
  Zhao, Chengqi Deng, Chenyu Zhang, Chong Ruan, and 1 others.
  2024{\natexlab{b}}.
\newblock Deepseek-v3 technical report.
\newblock \emph{arXiv preprint arXiv:2412.19437}.

\bibitem[{Paperno et~al.(2016)Paperno, Kruszewski, Lazaridou, Pham, Bernardi,
  Pezzelle, Baroni, Boleda, and Fern{\'a}ndez}]{paperno2016lambada}
Denis Paperno, Germ{\'a}n Kruszewski, Angeliki Lazaridou, Quan~Ngoc Pham,
  Raffaella Bernardi, Sandro Pezzelle, Marco Baroni, Gemma Boleda, and Raquel
  Fern{\'a}ndez. 2016.
\newblock The lambada dataset: Word prediction requiring a broad discourse
  context.
\newblock \emph{arXiv preprint arXiv:1606.06031}.

\bibitem[{Sap et~al.(2019)Sap, Rashkin, Chen, LeBras, and
  Choi}]{sap2019socialiqa}
Maarten Sap, Hannah Rashkin, Derek Chen, Ronan LeBras, and Yejin Choi. 2019.
\newblock Socialiqa: Commonsense reasoning about social interactions.
\newblock \emph{arXiv preprint arXiv:1904.09728}.

\bibitem[{Shazeer et~al.(2017)Shazeer, Mirhoseini, Maziarz, Davis, Le, Hinton,
  and Dean}]{shazeer2017outrageously}
Noam Shazeer, Azalia Mirhoseini, Krzysztof Maziarz, Andy Davis, Quoc Le,
  Geoffrey Hinton, and Jeff Dean. 2017.
\newblock Outrageously large neural networks: The sparsely-gated
  mixture-of-experts layer.
\newblock \emph{arXiv preprint arXiv:1701.06538}.

\bibitem[{Shoeybi et~al.(2019)Shoeybi, Patwary, Puri, LeGresley, Casper, and
  Catanzaro}]{shoeybi2019megatron}
Mohammad Shoeybi, Mostofa Patwary, Raul Puri, Patrick LeGresley, Jared Casper,
  and Bryan Catanzaro. 2019.
\newblock Megatron-lm: Training multi-billion parameter language models using
  model parallelism.
\newblock \emph{arXiv preprint arXiv:1909.08053}.

\bibitem[{Sun et~al.(2024)Sun, Liu, Luan, Gao, and
  Wang}]{sun-etal-2024-mixture}
Manxi Sun, Wei Liu, Jian Luan, Pengzhi Gao, and Bin Wang. 2024.
\newblock \href {https://doi.org/10.18653/v1/2024.emnlp-industry.118} {Mixture
  of diverse size experts}.
\newblock In \emph{Proceedings of the 2024 Conference on Empirical Methods in
  Natural Language Processing: Industry Track}, pages 1608--1621, Miami,
  Florida, US. Association for Computational Linguistics.

\bibitem[{Thompson et~al.(2020)Thompson, Greenewald, Lee, Manso
  et~al.}]{thompson2020computational}
Neil~C Thompson, Kristjan Greenewald, Keeheon Lee, Gabriel~F Manso, and 1
  others. 2020.
\newblock The computational limits of deep learning.
\newblock \emph{arXiv preprint arXiv:2007.05558}, 10.

\bibitem[{Touvron et~al.(2023)Touvron, Lavril, Izacard, Martinet, Lachaux,
  Lacroix, Rozi{\`e}re, Goyal, Hambro, Azhar et~al.}]{touvron2023llama}
Hugo Touvron, Thibaut Lavril, Gautier Izacard, Xavier Martinet, Marie-Anne
  Lachaux, Timoth{\'e}e Lacroix, Baptiste Rozi{\`e}re, Naman Goyal, Eric
  Hambro, Faisal Azhar, and 1 others. 2023.
\newblock Llama: Open and efficient foundation language models.
\newblock \emph{arXiv preprint arXiv:2302.13971}.

\bibitem[{Wang et~al.(2024)Wang, Sun, Xie, Li, Zhu, Yang, Zhao, Han, Kang, Wang
  et~al.}]{wang2024hmoe}
An~Wang, Xingwu Sun, Ruobing Xie, Shuaipeng Li, Jiaqi Zhu, Zhen Yang, Pinxue
  Zhao, JN~Han, Zhanhui Kang, Di~Wang, and 1 others. 2024.
\newblock Hmoe: Heterogeneous mixture of experts for language modeling.
\newblock \emph{arXiv preprint arXiv:2408.10681}.

\bibitem[{Wei et~al.(2022)Wei, Tay, Bommasani, Raffel, Zoph, Borgeaud,
  Yogatama, Bosma, Zhou, Metzler et~al.}]{wei2022emergent}
Jason Wei, Yi~Tay, Rishi Bommasani, Colin Raffel, Barret Zoph, Sebastian
  Borgeaud, Dani Yogatama, Maarten Bosma, Denny Zhou, Donald Metzler, and 1
  others. 2022.
\newblock Emergent abilities of large language models.
\newblock \emph{arXiv preprint arXiv:2206.07682}.

\end{thebibliography}

\clearpage

\appendix

\section{Appendix}
\label{sec: appendix}

\subsection{Background: Mixture of Experts}
An MoE layer typically includes the gating model $G_1(\cdot)\cdots G_N(\cdot)$, the expert networks $E_1(\cdot)\cdots E_N(\cdot)$, and the routing mechanism, where $N$ denotes the number of experts. The gating model serves as the mathematical implementation of a router, determining how input data is allocated to experts. Specifically, the gating model with learnable weights $W \in \mathbb{R}^{h_{input} \times h}$ selects the top $k$ experts and combines the outputs of these top $k$ experts to produce the output $y \in \mathbb{R}^h$, where $h_{input}$ is the dimension of input $x$ and $h$ is the dimension of the hidden layer. The output of an MoE layer can be expressed as,\par

\begin{small}
\begin{align}
    &y = \sum_{i=1}^{N}G_i(x)E_i(x)   \\
    G_i(x) &= Softmax(topK(H(x))) \\
    &H(X)_i = (x\cdot W)_i       \\
    TopK&(v, k)_i = 
    \begin{cases}
    v_i, & v_i \in topk(v)\\
    -\infty, & \text{otherwise}
    \end{cases}
\end{align}
\end{small}

\subsection{Experimental Setup}

\paragraph{Compute Infrastructure.}
All models were trained on a 16-node GPU cluster, with each node equipped with eight NVIDIA GPUs. We used the Megatron-LM framework (\citeauthor{shoeybi2019megatron}, \citeyear{shoeybi2019megatron}) to implement our MoHGE variants, as well as the dense and MoE baseline models.

\paragraph{Pretraining Data.}  
Our pretraining corpus was created by merging and deduplicating three large English datasets: DataComp-LM, FineWeb, and The Pile. The combined corpus underwent standard noise filtering and quality checks to ensure data integrity. For all experiments, we sampled 0.58 trillion tokens from this cleaned, unified corpus.

\paragraph{Model Configurations.}
We evaluated three Transformer variants at the 1B 3B, and 14B parameter scales: a Dense model whose parameters are equal to the active parameters of the MoE baseline, a uniform-expert MoE baseline, and our proposed MoHGE architecture with heterogeneous expert groups. The MoE baseline is adapted from DeepSeekV2 (\citeauthor{liu2024deepseekv2}, \citeyear{liu2024deepseekv3}), with hyperparameters adjusted to align parameter counts across models for fair comparison. Detailed architectural configurations for all evaluated models are summarized in \cref{tab:placeholder}.

\paragraph{Training Hyperparameters.}  
Each MoE model was trained for 2 full epochs on the 0.58 trillion–token corpus, using a fixed sequence length of 4,096. We used the AdamW optimizer with $\beta_1 = 0.9$, $\beta_2 = 0.95$, and a weight decay of 0.1. A cosine-decay learning rate schedule was applied, starting at $3 \times 10^{-4}$ and annealing to a minimum of $3 \times 10^{-5}$.

\begin{table}[!h]
\centering
\scriptsize
\setlength{\tabcolsep}{2pt}
\begin{tabular}{p{3cm}|c|c|c}
\hline
\textbf{Configuration} & \textbf{1B Scale} & \textbf{3B Scale}  & \textbf{14B Scale} \\
\hline \hline
\multicolumn{4}{l}{\textbf{Shared Configuration}} \\ \hline
Transformer Layers & 9     & 15      & 36 \\ \hline
Input Dim          & 1024  & 1024    & 1024 \\\hline
Attention Heads    & 16    & 16      & 16\\
\hline \hline 
\multicolumn{4}{l}{\textbf{Dense Model}} \\ \hline
FFN Hidden Dim & 4096    & 6144    & 8192  \\\hline
\hline \hline 
\multicolumn{4}{l}{\textbf{MoE Baseline}} \\ \hline
$N_e$  & 32     & 64      & 128   \\\hline
$K_e$  & 6      & 6       & 6      \\ \hline
Shared Experts $N_s$ & 2 & 2    & 2   \\ \hline
Expert Hidden Dim & 832 & 1024  & 1280\\ \hline
\hline
\multicolumn{4}{l}{\textbf{MoHGE}} \\ \hline
$N_g$    & 8         & 8        & 8 \\ \hline
$K_g$    & 3         & 3        & 3 \\ \hline
$N_e$    & 32        & 64       & 128 \\ \hline
$K_e$    & 6         & 6        & 6\\ \hline
Shared Experts $N_s$ & 2  & 2   & 2\\ \hline
\makecell[l]{Hidden Dims of \\ Expert Groups} & 
\makecell[l]{\{256, 320, \\ 384, 512, \\ 640, 768, \\ 832, 896\}} 
& \makecell[l]{\{384, 512, \\ 640, 768,\\ 896, 1024, \\ 1152, 1280\}} 
& \makecell[l]{\{640, 768, \\ 896, 1024,\\ 1152, 1280, \\ 1408, 1536\}} \\ \hline
\end{tabular}
\caption{Architecture configurations of the evaluated models at both 1B, 3B and 14B parameter scales.}
\label{tab:placeholder}
\end{table}

\begin{table*}[t]
\centering
\scriptsize
\setlength{\tabcolsep}{9.5pt}
\begin{tabular}{c|cc|c|ccccc}
\hline
Model               & $\alpha_{Exp}$ & $\alpha_{Grp}$  & \begin{tabular}[c]{@{}c@{}@{}}Activated \\ Parameters \\ of Experts\end{tabular} & MMLU & SIQA & PIQA & LAMBADA  & TriviaQA \\ \hline
\multirow{6}{*}{\begin{tabular}[c]{@{}c}MoHGE-1B\end{tabular}} 
                    & 0      & 0     & 139M    & 25.43     & 34.73    &  47.62   &52.20 & 25.03 \\
                    & 2.5e-3 & 0     &  132M   & 25.61    & 34.82    &  47.93 & 53.35 & 25.37  \\ 
                    & 5e-3   & 0     &  131M   &  25.87    & 34.74    &  48.77  & 53.14  & 25.20  \\ 
                    & 2.5e-3 & 1e-4  &  122M   &  \textbf{25.98}    & \textbf{35.17}    &  \underline{48.85}  & \textbf{53.75}  &  \textbf{25.42} \\ 
                    & 2.5e-3  & 1e-3 & 122M   &  25.94    & \underline{35.10}     &  48.28  & 52.99 & 25.25  \\ 
                    & 5e-3  & 1e-4   & 119M  &  \underline{25.96}    & 34.86    &  48.12  &53.16 & \underline{25.39} \\ \hline
\multirow{6}{*}{\begin{tabular}[c]{@{}c}MoHGE-3B\end{tabular}} 
                    & 0     & 0       & 324M    & 25.88    &  35.29    &  48.65    & 61.37     & 38.01\\ 
                    & 2.5e-3     & 0  & 307M  &  26.11    & 35.45     &  48.53   & 61.62   & 38.53\\ 
                    & 5e-3     & 0    & 310M   &  26.03        & 35.32     & 48.67    & 61.75   & 38.45\\ 
                    & 2.5e-3  & 1e-4  & 295M   &  \underline{26.41}     & \textbf{35.57}     & \textbf{49.08}     & \textbf{62.37}  & \textbf{39.20} \\ 
                    & 2.5e-3  & 1e-3  & 297M   &  \textbf{26.46}    &  35.12    &  48.83   & \underline{62.10}  & 38.68\\ 
                    & 5e-3  & 1e-4 & 289M     &  26.27    &  \underline{35.47}  &  48.21    & 61.85  &  38.51\\  \hline

\multirow{6}{*}{\begin{tabular}[c]{@{}c}MoHGE-14B\end{tabular}} 
                    & 0     & 0       & 897M    & 31.37    &  44.92    &  57.94    & 68.57     & 51.72 \\ 
                    & 2.5e-3     & 0  & 884M  &  31.18    & 45.03     &  58.07   & 68.95   & 51.86\\ 
                    & 5e-3     & 0    & 875M   &  \textbf{31.71}   & 45.39     & 58.27    & 68.90   & 52.29\\ 
                    & 2.5e-3  & 1e-4  & 843M   &  \underline{31.62}     & \textbf{45.62}     & \textbf{58.73}     & \textbf{69.89}  & \underline{52.49} \\ 
                    & 2.5e-3  & 1e-3  & 854M   &  30.87    &  45.07    &  58.22   & 69.10  & \textbf{52.75} \\ 
                    & 5e-3  & 1e-4 & 859M     &  31.38    &  44.78  &  58.15    & \underline{69.85}  &  51.67\\  \hline
\end{tabular}
\caption{The evaluation results for varying coefficients of the auxiliary loss function. The highest-performing score for each benchmark is highlighted in bold, while the second-highest score is underlined.}
\label{tab: cofficients}
\vspace{-1mm}
\end{table*}

\begin{table*}[!t]
\centering
\scriptsize
\setlength{\tabcolsep}{6.25pt}
\begin{tabular}{c|cc|ccccccc}
\hline
Method     & \begin{tabular}[c]{@{}c@{}}Total \\ Parameters\end{tabular} & \begin{tabular}[c]{@{}c@{}@{}}Activated \\ Parameters \\ of Experts\end{tabular} & MMLU  & SIQA  & GSM8K & LAMBADA & MATH & PIQA  & TriviaQA \\ \hline
Dense      & 0.570B                                                      & \textbf{\--- }                                                       & 25.41      &  34.93     &  1.81     &  51.87       &  1.22    &  44.85        &   25.05       \\ 
MoE-1B     & 1.098B     & 0.163B                                           & 25.38 & 35.12 & 1.74  & 53.20   & 1.26   & 46.09    & \textbf{25.86}    \\ 
MoHGE-1B  & 0.891B    & 0.122B                         
&  \textbf{25.98}     & \textbf{35.17} & \textbf{1.97}  & \textbf{53.75}   & \textbf{1.30} & \textbf{48.85} & 25.71    \\  \hline
\end{tabular}
\caption{Comparison between Dense model, MoE baseline and our MoHGE at 1B.}
\label{atab:evaluations}
\vspace{-1mm}
\end{table*}

\begin{table}[!htb]
\centering
\scriptsize
\scriptsize
\setlength{\tabcolsep}{10pt}
\begin{tabular}{l|ccc}
\hline
Benchmark & Dense-Model & MoE & MoHGE  \\ \hline
MMLU              & 6.22h     & 6.90h     & 6.77h   \\
SIQA              & 0.85h     & 0.93h     & 0.89h   \\
GSM8K             & 0.54h     & 0.59h     & 0.62h   \\
LAMBADA           & 2.10h     & 2.24h     & 2.22h   \\
MATH              & 2.12h     & 2.24h     & 2.21h   \\
PIQA              & 0.79h     & 0.85h     & 0.78h   \\
TriviaQA          & 3.87h     & 4.11h     & 3.95h   \\ \hline
\end{tabular}
\caption{The inference duration of the Dense, MoE and MoHGE models at 1B on downstream tasks.}
\label{atab: inference time}
\vspace{-1mm}
\end{table}

\subsection{Ablation Study on Auxiliary Loss Coefficients}

We conduct an ablation study to analyze the effect of different auxiliary loss coefficients on model performance. A coefficient of 0 indicates the absence of the auxiliary loss.

As shown in \cref{tab: cofficients}, the intra-group experts auxiliary loss yields a modest performance gain and setting $\alpha_{Exp}=2.5e-3$ achieves better results. Combining it with the group-wise auxiliary loss further improves results. Although the group-wise loss contributes only marginally to accuracy, it reduces the number of activated parameters. Based on the trade-off between evaluation performance and computational efficiency, we find that setting $\alpha_{Exp}=2.5e-3$ and $\alpha_{Grp}=1e-4$ enables the our models to achieve an optimal balance.





\subsection{Results at 1B Scale}

We also conducted training at the 1B scale, and the evaluation results and inference times obtained are shown in \cref{atab:evaluations} and \cref{atab: inference time}. It can be seen that our 1B-scale model has achieved consistent excellent performance and shorter inference time.



%

\end{document}